\documentclass[11pt,a4paper]{article}
\usepackage[hyperref]{acl2021}
\usepackage{times}
\usepackage{latexsym}
\usepackage{amsmath}
\usepackage{multirow}

\usepackage{enumitem} 
\usepackage{xfrac}

\usepackage{microtype}

\usepackage{graphicx}
\usepackage[utf8]{inputenc}
\usepackage[T1]{fontenc}
\usepackage[english]{babel}

\usepackage{microtype}

\aclfinalcopy % Uncomment this line for the final submission
\def\aclpaperid{2267} %  Enter the acl Paper ID here

\usepackage{booktabs, tabularx}
\usepackage{amssymb}
\usepackage{subfig}

\usepackage{todonotes}

\DeclareMathSymbol{\shortminus}{\mathbin}{AMSa}{"39}

\title{Analyzing the Source and Target Contributions\\ to Predictions in Neural Machine Translation}

  \author{Elena Voita$^{1,2}$\quad\quad Rico Sennrich$^{3,1}$\quad\quad Ivan Titov$^{1,2}$\bigskip\\
  $^1$University of Edinburgh, Scotland  \quad
  $^2$University of Amsterdam, Netherlands \\
  $^3$University of Zurich, Switzerland \\
  {\tt lena-voita@hotmail.com} \quad {\tt sennrich@cl.uzh.ch} \quad {\tt ititov@inf.ed.ac.uk}
}

\date{}

\begin{document}
\maketitle
\begin{abstract}

In Neural Machine Translation (and, more generally, conditional language modeling), the generation of a target token is influenced by two types of context: the source and the prefix of the target sequence. While many attempts to understand the internal workings of NMT models have been made, none of them explicitly evaluates relative source and target contributions to a generation decision. We argue that this relative contribution can be evaluated by adopting a variant of Layerwise Relevance Propagation~(LRP). Its underlying `conservation principle' makes relevance propagation unique: differently from other methods, it evaluates not an abstract quantity reflecting token importance, but the proportion of each token's influence. We extend LRP to the Transformer and conduct an analysis of NMT models which explicitly evaluates the source and target relative contributions to the generation process. We analyze changes in these contributions when conditioning on different types of prefixes, when varying the training objective or the amount of training data, and during the training process. We find that models trained with more data tend to rely on source information more and to have more sharp token contributions; the training process is non-monotonic with several stages of different nature.\footnote{We release the code at \url{https://github.com/lena-voita/the-story-of-heads}.}

\end{abstract}

\section{Introduction}

With the success of neural approaches to natural language processing, analysis of NLP models has become an important and active topic of research. In NMT, approaches to analysis include probing for linguistic structure~\cite{belinkov-etal-2017-neural,conneau-etal-2018-cram},
evaluating via contrastive translation pairs~\cite{sennrich-2017-grammatical,burlot-yvon-2017-evaluating,rios-gonzales-etal-2017-improving,tang-etal-2018-self},
inspecting model components, such as attention~\cite{monz-attention,voita18,tang-sennrich-nivre:2018:WMT,raganato-tiedemann:2018:BlackboxNLP,voita-etal-2019-analyzing} or neurons~\cite{dalvi2019one,bau2019neurons-in-mt}, among others. 

Unfortunately, although a lot of work on model analysis has been done, a question of how the NMT predictions are formed remains largely open. Namely, the generation of a target token is defined by two types of context, source and target, but there is no method which explicitly evaluates the relative contribution of source and target to a given prediction.
The ability to measure this relative contribution 
is important for model understanding since
previous work showed that NMT models often fail to effectively control information flow from source and target contexts. For example, adding context gates to dynamically control the influence of source and target leads to improvement for both RNN~\cite{tu-etal-2017-context,wang-etal-2018-neural-machine} and Transfomer~\cite{li2019regularized} models.
A more popular example is a model's tendency to generate hallucinations (fluent but inadequate translations); it is usually attributed to the inappropriately strong influence of target context. Several works observed that, when hallucinating, a model fails to properly use source: it produces a deficient attention matrix, where almost all the probability mass is concentrated on uninformative source tokens (EOS and punctuation)~\cite{lee2018hallucinations,berard-etal-2019-naver}.

We argue that a natural way to estimate how the source and target contexts contribute to generation is to apply 
Layerwise Relevance Propagation~(LRP)~\cite{bach2015pixel} to NMT models. LRP redistributes the information used for a prediction between all input elements keeping the total contribution constant. This `conservation principle' makes relevance propagation unique: differently from other methods estimating influence of individual tokens~\cite{alvarez-melis-jaakkola-2017-causal,he-etal-2019-towards,ma2018analysis}, LRP evaluates not an abstract quantity reflecting a token importance, but the proportion of each token's influence.

We extend one of the LRP variants to the Transformer and conduct the first analysis of NMT models which explicitly evaluates the source and target relative contributions to the generation process. We analyze changes in these contributions when conditioning on different types of prefixes (reference, generated by a model or random translations), when varying training objective or the amount of training data, and during the training process. We show that models suffering from exposure bias are more prone to over-relying on target history (and hence to hallucinating) than the ones where the exposure bias is mitigated. When comparing models trained with different amount of data, we find that extra training data teaches a model to rely on source information more heavily and to be more confident in the choice of important tokens. When analyzing the training process, we find that changes in training are non-monotonic and form several distinct stages (e.g., stages changing direction from decreasing influence of source to increasing).

Our key contributions are as follows:
\begin{itemize}
    \item we show how to use LRP to evaluate the relative contribution of source and target to NMT predictions;
    \item we analyze how the contribution of source and target changes when conditioning on different types of prefixes: reference, generated by a model or random translations;
    \item we show that models suffering from exposure bias are more prone to over-relying on target history (and hence to hallucinating);
    \item we find that (i)~with more data, models rely on source information more and have more sharp token contributions, (ii) the training process is non-monotonic with several distinct stages. 
    %\item we find that (i) models suffering from exposure bias are more prone to over-relying on target history (and hence to hallucinating), (ii)~with more data, models rely on source information more and have more sharp token contributions, (iii) the training process is non-monotonic with several distinct stages. 
  
\end{itemize}

\section{Layer-wise Relevance Propagation}
\label{sect:lrp}

Layer-wise relevance propagation is a framework which  decomposes the prediction 
of a deep neural network computed over an instance, e.g. an image or sentence, into relevance scores for  single input dimensions of the sample such as subpixels of an image or neurons of input token embeddings. The original LRP version was developed for computer vision models~\cite{bach2015pixel} and is not directly applicable to the Transformer (e.g., to the attention layers). In this section, we 
explain the general idea behind LRP, specify which of the existing LRP variants we use, and show how to extend LRP to the NMT Transformer model.\footnote{Previous work applying one of the LRP variants to NMT~\cite{lrp-ding-2017,voita-etal-2019-analyzing} do not describe extensions beyond the original LRP rules~\cite{bach2015pixel}.} 

\subsection{General Idea: Conservation Principle}

In its general form, 
LRP assumes that the model can be decomposed into several layers of computation. 
The first layer are the inputs (for example, the pixels of an image or tokens of a sentence), the last layer is the real-valued prediction output of the model $f$. The $l$-th layer is modeled as a vector $x^{(l)}=(x_{i}^{(l)})_{i=1}^{V(l)}$ with dimensionality $V(l)$. Layer-wise relevance propagation assumes that we have a relevance score $R_i^{(l+1)}$ for each dimension $x_{i}^{(l+1)}$ of the vector $x$ at layer $l + 1$. The idea is to find a relevance score  $R_i^{(l)}$ for each dimension  $x_i^{(l)}$ of the previous layer $l$ such that the following holds:
\begin{equation}
%f\!=\!\dots\!=\!\sum\limits_{i=1}^{V(l+1)}\!R_i^{(l+1)}\!=\sum\limits_{i=1}^{V(l)}\!R_i^{(l)}=\dots=\sum\limits_{i=1}^{V(1)}R_i^{(1)}.
f\!=\!...\!=\!\!\sum\limits_{i}\!R_i^{(l+1)}\!=\!\!\sum\limits_{i}\!R_i^{(l)}\!=\!...\!=\!\!\sum\limits_{i}\!R_i^{(1)}\!.
\vspace{-1ex}
\end{equation}

This equation represents a \textit{conservation principle}, 
which LRP exploits to back-propagate the prediction. Intuitively, this means that the total contribution of neurons at each layer is constant.

\subsection{Redistribution Rules}

Assume that we know the relevance $R_j^{(l\!+\!1)}$ of a neuron $j$ at network layer 
$l\!+\!1$ for the prediction~$f(x)$. 
Then we would like to decompose this relevance into messages $R_{i\leftarrow j}^{(l,l+1)}$ sent from the neuron~$j$ at layer $l+1$ to each of its input neurons~$i$ at layer~$l$.
For the conservation principle to hold, these messages $R_{i\leftarrow j}^{(l,l+1)}$ have to satisfy the %following 
constraint:
\vspace{-1ex}
\begin{equation}
R_j^{(l+1)} = \sum\limits_{i}R_{i\leftarrow j}^{(l,l+1)}.
\label{eq:lrp_inp_to_out}
\vspace{-1ex}
\end{equation}
Then we can define the relevance of a neuron $i$ at layer $l$ by summing all messages from neurons at layer $(l+1)$:
\vspace{-2ex}
\begin{equation}
R_i^{(l)} = \sum\limits_{j}R_{i\leftarrow j}^{(l,l+1)}.
\label{eq:lrp_out_to_inp}
\vspace{-1ex}
\end{equation}
Equations~(\ref{eq:lrp_inp_to_out}) and (\ref{eq:lrp_out_to_inp}) define the propagation of relevance from layer $l+1$ to layer $l$. 
The only thing that is missing is specific formulas for computing the messages $R_{i\leftarrow j}^{(l,l+1)}$. Usually, the message $R_{i\leftarrow j}^{(l,l+1)}$ has the following structure:
\begin{equation}
\vspace{-1ex}
R_{i\leftarrow j}^{(l,l+1)} = v_{ij}R_{j}^{(l+1)}, \ \ \ \ \sum\limits_{i}v_{ij}=1.
\label{eq:lrp_structure_of_message}
\end{equation}
Several versions of LRP satisfying equation~(\ref{eq:lrp_structure_of_message}) (and, therefore, the conservation principle) have been introduced: LRP-$\varepsilon$, LRP-$\alpha\beta$ and LRP-$\gamma$~\cite{bach2015pixel,Binder_2016,montavon2019layer}. 
We use LRP-$\alpha\beta$~\cite{bach2015pixel,Binder_2016}, which defines relevances at each step in such a way that they are positive. 

\paragraph{Rule for relevance propagation: the $\alpha\beta$-rule.}
Let us consider the simplest case of linear layers with non-linear activation functions, namely
%\vspace{-0.7ex}
\begin{equation*}
z_{ij} = x_i^{(l)}\!w_{ij}, \ \ z_j =\!\! \sum\limits_{i}\!\! z_{ij} + b_i, \ \ x_j^{(l+1)}\!\!= g(z_j), 
\vspace{-2ex}
\end{equation*}
where $w_{ij}$ is a weight connecting the neuron $x_i^{(l)}$ to neuron $x_j^{(l+1)}$, $b_j$ is a bias term, and $g$ is a non-linear activation function. Let 
\begin{equation*}
z_j^{+} = \sum\limits_i z_{ij}^{+} + b_j^{+}, \ \ \ \ z_j^{-} = \sum\limits_i z_{ij}^{-} + b_j^{-},
\vspace{-2ex}
\end{equation*}
where $\square^{+}=\max(0, \square)$ and $\square^{-}=\min(0, \square)$.
Then the $\alpha\beta$-rule~\cite{bach2015pixel,Binder_2016} is given by the equation
\begin{equation}
R_{i\leftarrow j}^{(l,l+1)} = R_{j}^{(l+1)} \cdot \left(\alpha\cdot\frac{z_{ij}^{+}}{z_j^{+}} + \beta\cdot\frac{z_{ij}^{-}}{z_j^{-}}\right),
\label{eq:alpha_beta_rule}
%\vspace{-1ex}
\end{equation}
where $\alpha+\beta=1$. Note that all terms in the brackets are always positive: negative signs of $z_j^{-}$ and $z_{ij}^{-}$ cancel out when evaluating the ratio.

This propagation method allows to control manually the importance of positive and negative evidence by choosing different $\alpha$ and $\beta$. For example, $\alpha, \beta=\frac12$ treats positive and negative contributions as equally important, while $\alpha=1$, $\beta=0$ considers only positive contributions. In our experiments, we use  $\alpha=1$, $\beta=0$.

Note that~(\ref{eq:alpha_beta_rule}) is directly applicable to all layers for which there exist  functions $g_j$ and $h_{ij}$ such that
\begin{equation}
x_j^{(l+1)} = g_j\left(\sum\limits_ih_{ij}(x_{i}^{(l)})\right).
\label{eq:base_allowed_structure}
\end{equation}
These layers include linear, convolutional and max-pooling operations. Additionally, pointwise monotonic activation functions $g_j$ (e.g., ReLU) are ignored by LRP~\cite{bach2015pixel}.

\paragraph{Propagating relevance through attention layers.}
For the structures that do not fit  the form~(\ref{eq:base_allowed_structure}) (e.g., softmax operation, layer normalization~\cite{ba2016layer}, residual connections), the weighting $v_{ij}$ can be obtained by performing a first order Taylor expansion of a neuron~$x_j^{(l+1)}$~\cite{bach2015pixel,Binder_2016}. 

For attention layers in the Transformer, we extend the approach by~\citet{Binder_2016}. Namely, let $x_j^{(l+1)} = f(x^{(l)})$, $f(x)=f(x_1, \dots, x_n)$. Then by Taylor expansion at some  point $\hat{x}=(\hat{x}_1, \dots, \hat{x}_n)$, we get 
\begin{equation*}
f(\hat{x}) \approx f(x^{(l)}) + \sum\limits_{i\leftarrow j}\frac{\partial f}{\partial x_i}(x^{(l)}) \cdot (\hat{x}_i - x^{(l)}_i),
\end{equation*}
\begin{equation*}
x_j^{(l+1)}\!\!=\!f(x^{(l)}) \approx f(\hat{x}) + \sum\limits_{i\leftarrow j}\frac{\partial f}{\partial x_i}(x^{(l)}) \cdot (x^{(l)}_i-\hat{x}_i).
\end{equation*}
Elements of the sum can be assigned to incoming neurons, and the zero-order term can be redistributed equally between them. This leads to the following decomposition:
\begin{equation}
z_{ij} = \frac{1}{n}f(\hat{x}) + \frac{\partial f}{\partial x_i}(x^{(l)}) \cdot (x^{(l)}_i-\hat{x}_i).
\label{eq:decomposition_attn_lrp}
\end{equation}
We use the zero vector in place of $\hat{x}$. Equation~(\ref{eq:decomposition_attn_lrp}), along with the standard redistribution rules~(\ref{eq:alpha_beta_rule}), defines relevance propagation for complex non-linear layers. In the Transformer, we apply equation~(\ref{eq:decomposition_attn_lrp}) to the softmax operations in the attention layers; all other operations inside the attention layers are linear functions, and the rule~(\ref{eq:alpha_beta_rule}) can be used. We also apply equation~(\ref{eq:decomposition_attn_lrp}) to layer normalization operations and residual connections.

\subsection{LRP for Conditional Language Models}
\label{sect:lrp_for_transformer}

Given a source sequence $x = (x_1, \dots, x_S)$ and a target sequence $y=(y_1, \dots, y_T)$, standard autoregressive NMT models (or, in a more broad sense, conditional language models) are trained to predict words in the target sequence, word by word. Formally, at each generation step such models predict $p(y_t|x_{1:S}, y_{1:t-1})$ relying on both source tokens $x_{1:S}$ and already generated target tokens $y_{1:t-1}$. Using LRP, we evaluate relative contribution of all tokens, source and target, to the current prediction. 

\paragraph{Propagating through decoder and encoder.} At first glance, it can be unclear how to apply a layer-wise method to a not completely layered architecture (such as encoder-decoder). This, however, is rather straightforward and is done in two steps: 
\begin{enumerate}
    \item total relevance is propagated through the decoder. Since the decoder uses representations from the final encoder layer, part of the relevance `leaks' to the encoder; this happens at each decoder layer;
    \item relevance leaked to the encoder is propagated through the encoder layers.
\end{enumerate}
The total contribution of neurons in each decoder layer is not preserved (part of the relevance leaks to the encoder), but the total contribution of all tokens~-- across the source and the target prefix~-- remains equal to the model prediction.

We evaluate relevance of input neurons to the top-1 logit predicted by a model. Then token relevance (or its contribution) is the sum of relevances of its neurons. 

\paragraph{Notation.} Without loss of generality, we can assume that the total relevance for each prediction equals 1.\footnote{More formally, if we evaluate relevance for top-1 logit predicted by a model, then the total relevance is equal to the value of this logit. However, the conservation principle allows us to assume that this logit is equal to 1 and to consider relative contributions.}
Let us denote by $\textsc{r}_t(x_i)$ and $\textsc{r}_t(y_j)$ the contribution of source token $x_i$ and target token $y_j$ to the prediction at generation step $t$, respectively. Then source and target contributions are defined as
$\textsc{r}_t(\text{source})=\sum\limits_{i}\textsc{r}_t(x_i)$, \ \ 
$\textsc{r}_t(\text{target})=\sum\limits_{j=1}^{t-1}\textsc{r}_t(y_j)$.

Note that \ \ \ $\forall~t \ \ \textsc{r}_t(\text{source})\!+\!\textsc{r}_t(\text{target})\!=\!1;$\\
$\textsc{r}_1(\text{source})\!=\!1, \ \textsc{r}_1(\text{target})\!=\!0$,  and $\forall j~\geq~t~~\textsc{r}_t(y_j)\!=\!0.$

\section{Experimental setting}

\paragraph{Model.} We follow the setup of Transformer base model~\cite{attention-is-all-you-need} with the standard training setting. More details on hyperparameters and the optimizer can be found in the appendix.

\paragraph{Data.} We use random subsets of the WMT14 En-Fr dataset of different size: 1m, 2{.}5m, 5m, 10m, 20m, 30m sentence pairs. In Sections~\ref{sect:getting_acquainted} and~\ref{sect:training_stages}, we report results for the model trained on the 1m subset. In Section~\ref{sect:data_amount}, we show how the results depend on the amount of training data.

\paragraph{Evaluating LRP.} The $\alpha\beta$-LRP we use requires choosing values for $\alpha$ and $\beta$, $\alpha + \beta=1$. We consider only positive contributions, i.e. we choose $\alpha=1$, $\beta=0$.\footnote{In preliminary experiments, we performed visual inspection of the contribution heatmaps and observed that $\alpha, \beta=\frac12$ (i.e., using also negative contributions) do not lead to reasonable contribution patterns.}

\paragraph{Reporting results.} All presented results are averaged over an evaluation dataset of 1000 sentence pairs. In each evaluation dataset, all examples have the same number of tokens in the source, as well as in the target (e.g., 20 source and 23 target tokens; the exact number for each experiment is clear from the results).\footnote{Note that we have to fix the number of tokens in the source and target to get reliable comparisons. We choose sentences of length 20 and 23 because these are among the most frequent sentence lengths in the dataset. We also looked at sentences with 16, 25, 29 tokens -- observed patterns were the same.}

\section{Getting Acquainted}
\label{sect:getting_acquainted}

In this section, we explain general patterns in  model behavior and illustrate the usage of LRP by evaluating different statistics within a single model. Later, we will show how these results change when varying the amount of training data~(Section~\ref{sect:data_amount}) and during model training~(Section~\ref{sect:training_stages}). 

\begin{figure}[t!]
    \centering
    \subfloat[]
    {\includegraphics[scale=0.21]{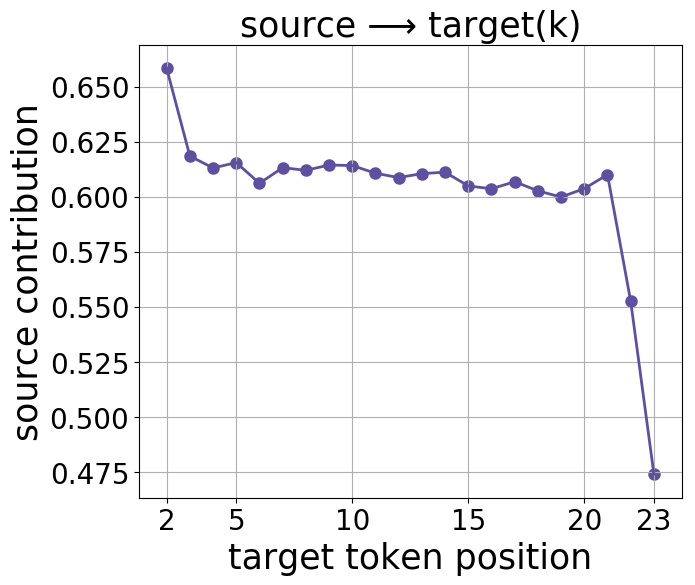}}
    \quad
    \subfloat[]
    {\includegraphics[scale=0.21]{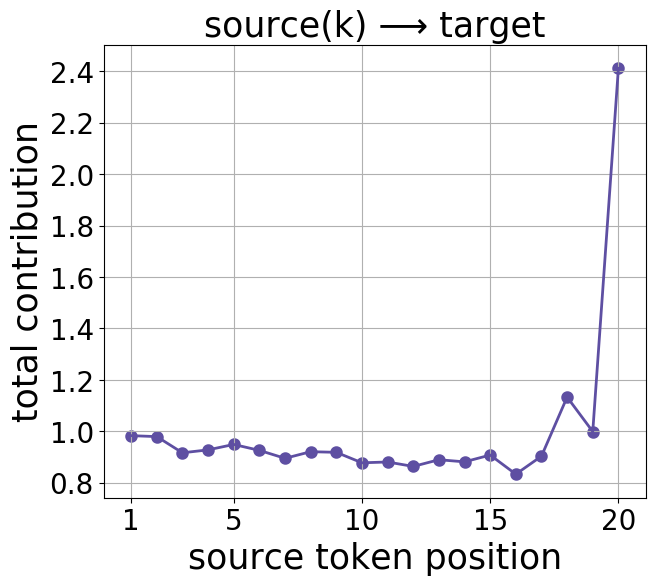}}
    \vspace{-1ex}
   \caption{(a) contribution of the whole source at each generation step; (b) total contribution of source tokens at each position to the whole target sentence.}
   \vspace{-2ex}
    \label{fig:general_graphs_contribution}
\end{figure}

\subsection{Changes in contributions}
\label{sect:getting_acquainted_amount}

Here we evaluate changes in the source contribution during generation, and in contributions of source tokens at different positions to entire output.

\paragraph{Source $\longrightarrow$ target(k).} For each generation step $t$, we evaluate total contribution of source $\textsc{r}_t(\text{source})$. Note that this is equivalent to evaluating total contribution of prefix since $\textsc{r}_t(\text{prefix})=1-\textsc{r}_t(\text{source})$ (Section~\ref{sect:lrp_for_transformer}).

Results are shown in
Figure~\ref{fig:general_graphs_contribution}(a).\footnote{Since the first token is always generated solely relying on the source, we plot starting from the second token.} We see that, during the generation process, the influence of source decreases (or, equivalently, the influence of the prefix increases). This is expected: with a longer prefix, the model has less uncertainty in deciding which source tokens to use, but needs to control more for fluency. 
There is also a large drop of source influence for the last token: apparently, to generate the EOS token, the model relies on prefix much more than when generating other tokens.

\paragraph{Source(k) $\longrightarrow$ target.} Now we want to understand if there is a tendency to use source tokens at certain positions more than tokens at the others. For each source token position $k$, we evaluate its total contribution to the whole target sequence. 
To eliminate the effect of decreasing source influence during generation, at each step $t$ we normalize source contributions $\textsc{r}_t(x_k)$ over the total contribution of source at this step $\textsc{r}_t(\text{source})$. Formally, for the $k$-th  token we evaluate $\sum\limits_{t=1}^T\textsc{r}_t(x_k)/\textsc{r}_t(\text{source})$. For convenience, we multiply the result by $\frac{S}{T}$: this makes the average total contribution of each token equal to 1.

Figure~\ref{fig:general_graphs_contribution}(b) shows that the end-of-sentence token is used much more actively than the rest; the two tokens before that (i.e., the final punctuation mark and the last content token) also influence translations more than other tokens. We hypothesize that this is because these last tokens are relevant not only for generation of the corresponding target tokens, but also for earlier tokens. For example, they may be used to account for the distance to the end of sentence or to understand the tone for the whole sentence. Except for these last three tokens, source tokens at earlier positions influence translations more than tokens at later ones. This may be because the alignment between English and French languages is roughly monotonic. We leave for future work investigating the changes in this behavior for language pairs with more complex alignment (e.g., English-Japanese).

\begin{figure}[t!]
    \centering
    \subfloat[]
    {\includegraphics[scale=0.21]{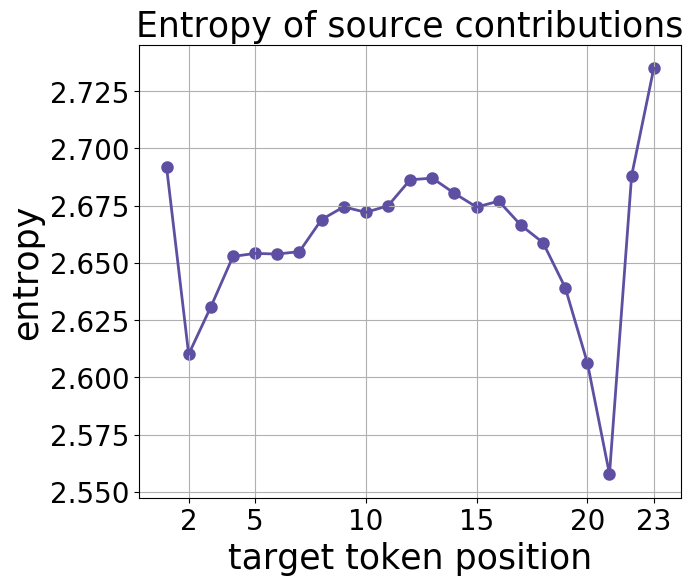}}
    \quad
    \subfloat[]
    {\includegraphics[scale=0.21]{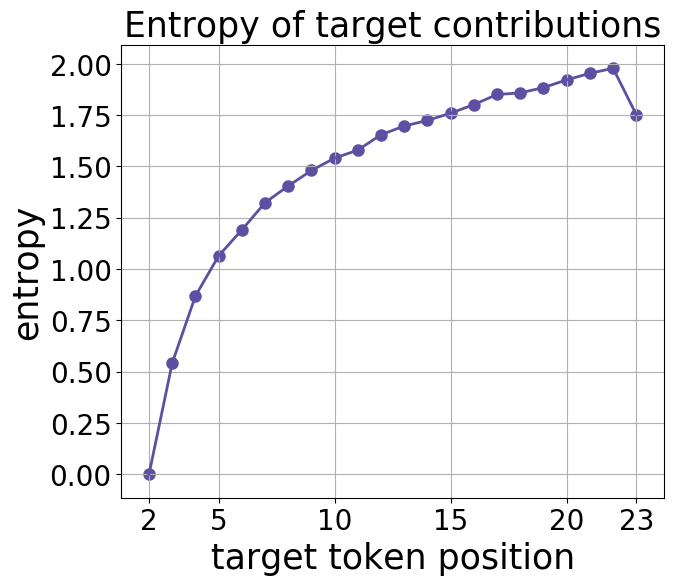}}
    \vspace{-1ex}
   \caption{For each generation step, the figure shows entropy of (a)~source, (b)~target contributions.}
   \vspace{-2ex}
    \label{fig:general_graphs_entropy}
\end{figure}

\subsection{Entropy of contributions}
\label{sect:getting_acquainted_entropy}
Now let us look at how `sharp' contributions of source or target tokens are at different generation steps. For each step~$t$, we evaluate entropy of (normalized) source or target contributions:
$\left\{\!\textsc{r}_t(x_i)/\textsc{r}_t(\text{source})\!\right\}_{i=1}^{S}$ or $\left\{\!\textsc{r}_t(y_j)/\textsc{r}_t(\text{target})\!\right\}_{j=1}^{t-1}$.

\paragraph{Entropy of source contributions.} Figure~\ref{fig:general_graphs_entropy}(a) shows that during generation, entropy increases until approximately $2/3$ of the translation is generated, then decreases when generating the remaining part. Interestingly, for the last punctuation mark and the EOS token, entropy of source contributions is very high: the decision to complete the sentence requires broader context.

\paragraph{Entropy of target contributions.} Figure~\ref{fig:general_graphs_entropy}(b) shows that entropy of target contributions is higher for longer prefixes. This means that the model does use longer contexts in a non-trivial way.

\subsection{Reference, Model and Random Prefixes}
\label{sect:getting_acquainted_prefixes}

\begin{figure}[t!]
    \centering
    \subfloat[]
    {\ \ \includegraphics[scale=0.195]{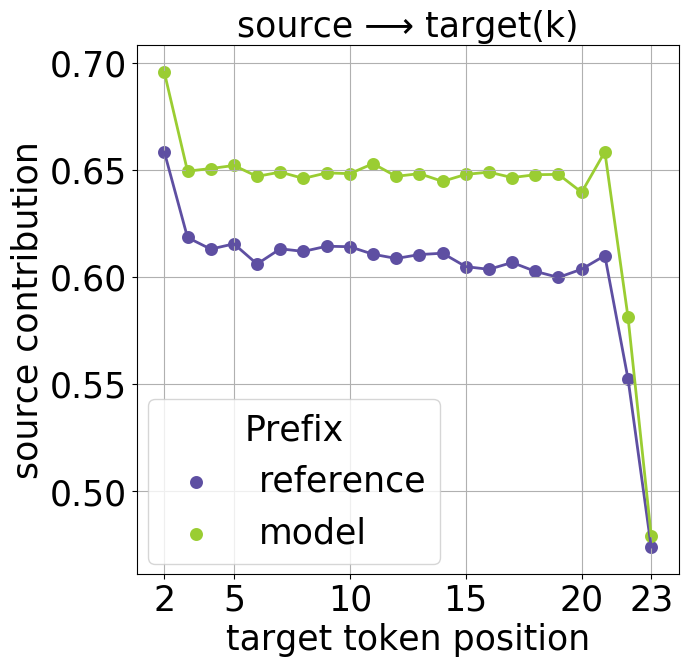}}
    \quad
    \subfloat[]
    {\includegraphics[scale=0.195]{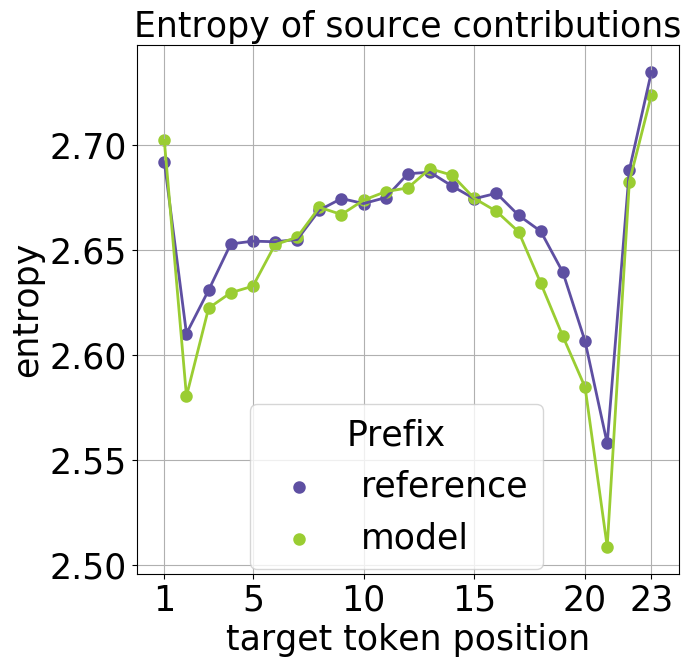}}
    \vspace{-1ex}
    \\
    \subfloat[]
    {\includegraphics[scale=0.195]{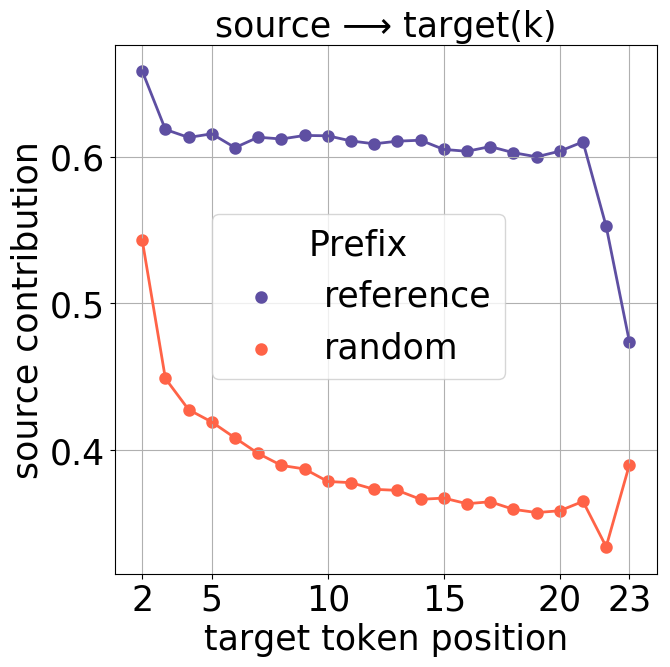}}
    \quad
    \subfloat[]
    {\includegraphics[scale=0.195]{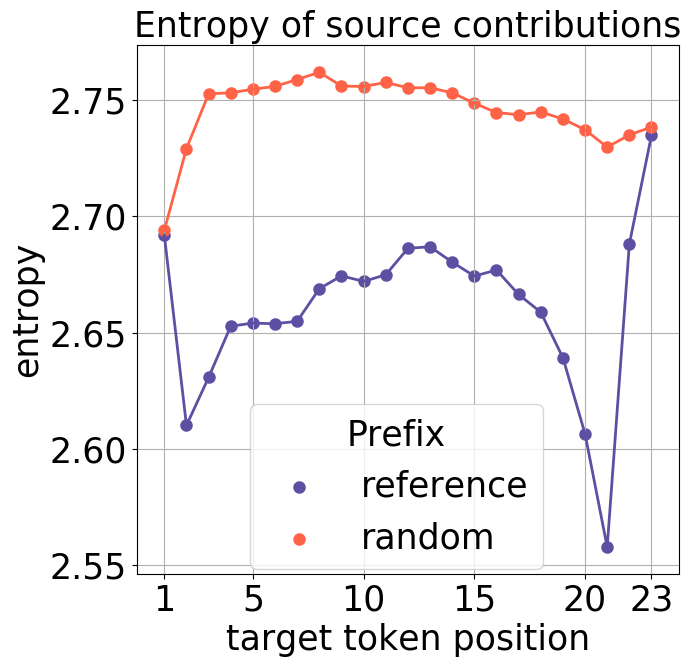}}
    \vspace{-1ex}
  \caption{(a,~c) contribution of source, (b,~d) entropy of source contributions.}
  \vspace{-2ex}
    \label{fig:prefix}
\end{figure}

Let us now look at how  model behavior changes when feeding different types of prefixes: prefixes of reference translations, translations generated by the model, and random sentences in the target language.\footnote{Random prefixes come from the same evaluation set, but with shuffled target sentences.} As in previous experiments, we evaluate relevance for top-1 logit predicted by the model. 

\paragraph{Reference vs model prefixes.} When feeding model-generated prefixes, the model uses source more (Figure~\ref{fig:prefix}(a)) and has more focused source contributions (lower entropy in Figure~\ref{fig:prefix}(b)) than when generating the reference. This may be because model-generated translations are `easier' than references. For example, beam search translations contain fewer rare tokens~\cite{burlot-yvon-2018-using,pmlr-v80-ott18a}, are simpler syntactically~\cite{burlot-yvon-2018-using} and, according to the fuzzy reordering score~\cite{talbot-etal-2011-lightweight}, model translations have significantly less reordering compared to the real parallel sentences~\cite{zhou2019understanding}. 
As we see from our experiments, these simpler model-generated prefixes allow for the model to rely on the source more and to be more confident when choosing relevant source tokens.

\paragraph{Reference vs random prefixes.} Results for random sentence prefixes are given in Figures~\ref{fig:prefix}c,~\ref{fig:prefix}d. 
The reaction to random prefixes helps us study the self-recovery ability of NMT models.
Previous work has found that models can fall into a hallucination mode where ``the decoder ignores context from the encoder and samples from its language mode'' \citep{koehn-knowles-2017-six,lee2018hallucinations}.
In contrast,  \citet{he2019quantifying} found that a language model is able to recover from artificially distorted history input and generate reasonable samples.

%Our results show evidence for both. At the beginning of the generation process, the model tends to rely more on the source context when given a random prefix compared to the reference prefix, indicating a self-recovery mode. However, when the prefix becomes longer, the model choice shifts towards ignoring the source and relying more on the target: Figure~\ref{fig:prefix}c shows a large drop of source influence for later positions.

Our results show that the model tends to fall into hallucination mode even when a random prefix is very short, e.g.\ one token: we see a large drop of source influence for all positions (Figure~\ref{fig:prefix}c).
 Figure~\ref{fig:prefix}d also shows that with a random prefix, the entropy of source contributions is high and is roughly constant.

\section{Exposure Bias and Source Contributions}
\label{sect:hallucinations}

The results in the previous section %, e.g. with random prefixes, 
agree with some observations made in previous work studying self-recovery and hallucinations. In this section, we illustrate more explicitly how our methodology can be used to shed light on the effects of exposure bias and training objectives.

\citet{wang-sennrich-2020-exposure} empirically link the hallucination mode to exposure bias \citep{ranzato2015sequence}, i.e.\ the mismatch between the gold history seen at training time, and the (potentially erroneous) model-generated prefixes at test time.
The authors hypothesize that exposure bias leads to an over-reliance on target history, and show that Minimum Risk Training (MRT), which does not suffer from exposure bias, reduces hallucinations. However, they did not directly measure this over-reliance on target history.
Our method is able to directly test whether there is indeed an over-reliance on the target history with MLE-trained models, and more robust inclusion of source context with MRT.
We also consider a simpler heuristic, word dropout, which we hypothesize to have a similar effect.
%Table~\ref{tab:training_regimes} summarizes the properties of the models.

% \begin{table}[t!]
% \centering
% \begin{tabular}{ccccc}
% \toprule

% & \bf Base & \multicolumn{2}{c}{\!\bf Word dropout\!} & \bf MRT\!\!\! \\
% &  &  source & \!\! target &  \\
%  %& \bf base &\!\! \bf wd, src & \!\!\bf wd, tgt\!\! & \bf MRT\!\!\! \\
% \midrule
% \multicolumn{5}{l}{\!\!\!\!\textbf{Loss} }\\
% sequence-level\! & no & no & no & yes \\
% \midrule
% \multicolumn{5}{l}{\!\!\!\!\textbf{Training prefixes} }\\
% not only gold\! & no & no & yes & yes \\
% \bottomrule
% \end{tabular}
% \caption{Properties of the training regimes.}
% \label{tab:training_regimes}
% \end{table}

\paragraph{Minimum Risk Training}~\cite{shen-etal-2016-minimum} is a sentence-level objective that inherently avoids exposure bias. It minimises the expected loss (`risk') with respect to the posterior distribution:
\vspace{-1ex}
\begin{equation*}
\mathcal{R}(\theta) = \sum\limits_{(x, y)}\sum\limits_{\Tilde{y}\in\mathcal{Y}(x)}P(\Tilde{y}|x, \theta)\Delta(\Tilde{y}, y),
\end{equation*}
where $\mathcal{Y}(x)$ is a set of candidate translations for~$x$,  $\Delta(\Tilde{y}, y)$ is the discrepancy between the model prediction~$\Tilde{y}$ and the gold translation~$y$ (e.g., a negative smoothed sentence-level BLEU). 
More details on the method can be found in~\citet{shen-etal-2016-minimum} or~\citet{edunov-etal-2018-classical}; training details for our models are in the appendix.

\paragraph{Word Dropout} is a simple data augmentation technique. During training, it replaces some of the tokens with a special token (e.g., UNK) or a random token (in our experiments, we replace $10\%$ of the tokens with random). When used on the target side, it may serve as the simplest way to alleviate exposure bias: it exposes a model to something other than gold prefixes. This is not true when used on the source side, but for analysis, we consider both variants.

% \begin{figure}[t!]
%     \centering
%     \subfloat[]
%     {\includegraphics[scale=0.20]{pict/random_prefix_source_mrt_wd.png}}
%     \quad
%     \subfloat[]
%     {\includegraphics[scale=0.20]{pict/random_prefix_entropy_mrt_wd.png}}
%     \vspace{-1ex}
%   \caption{Results for random prefixes. For each generation step, the figure shows (a)~contribution of source, (b)~entropy of source contributions.}
%   \vspace{-2ex}
%     \label{fig:mrt_wd_random_prefix}
% \end{figure}

\subsection{Experiments}

\begin{figure}[t!]
    \centering
    \subfloat[]
    {\includegraphics[scale=0.20]{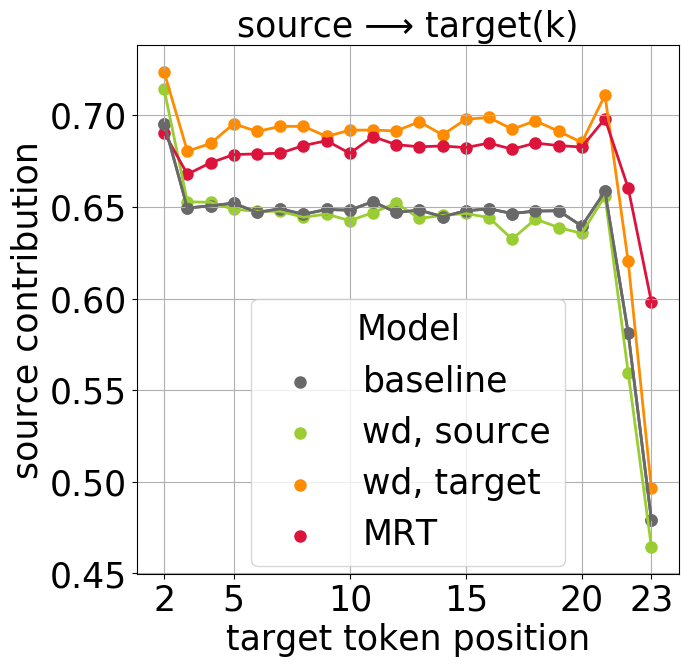}}
    \quad
    \subfloat[]
    {\includegraphics[scale=0.20]{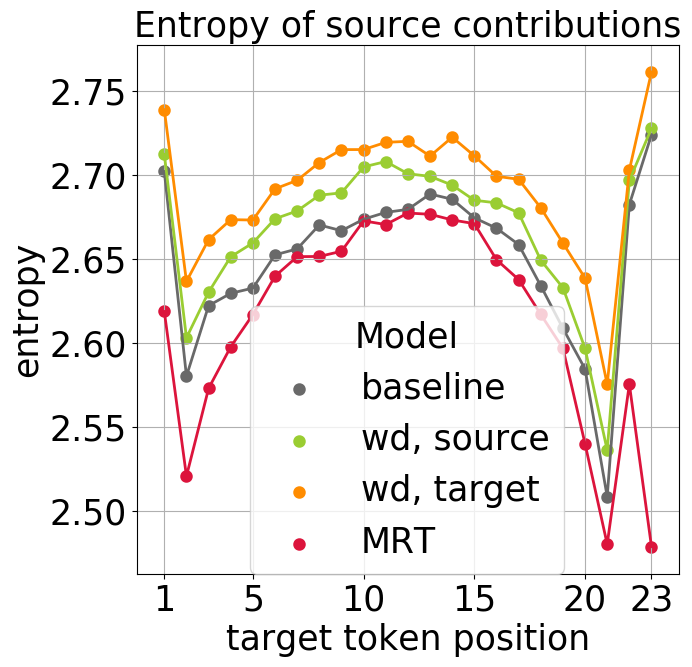}}
    \vspace{-1ex}
  \caption{Contribution of source (a) and entropy of source contributions (b) with model-generated prefixes. %For each generation step, the figure shows (a)~contribution of source, (b)~entropy of source contributions.
  }
  \vspace{-2ex}
    \label{fig:mrt_wd_model_prefix}
\end{figure}

\begin{figure}[t!]
    \centering
    \subfloat[]
    {\includegraphics[scale=0.20]{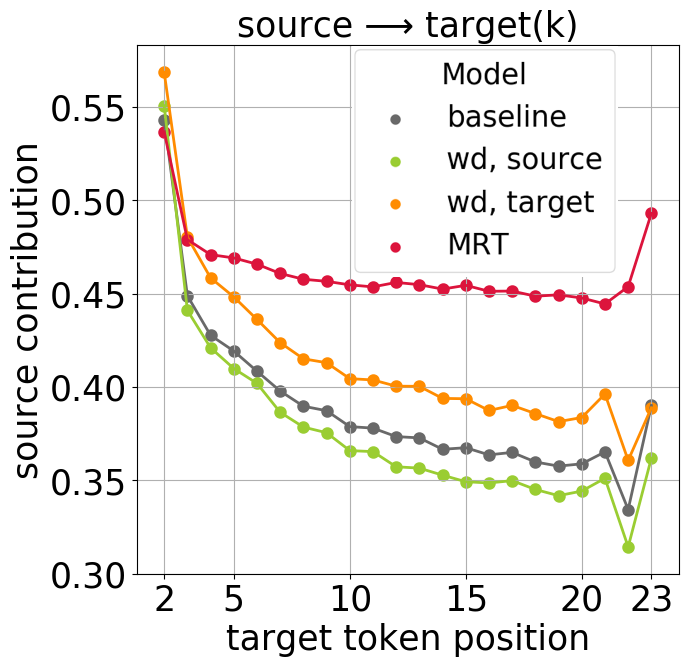}}
    \quad
    \subfloat[]
    {\includegraphics[scale=0.20]{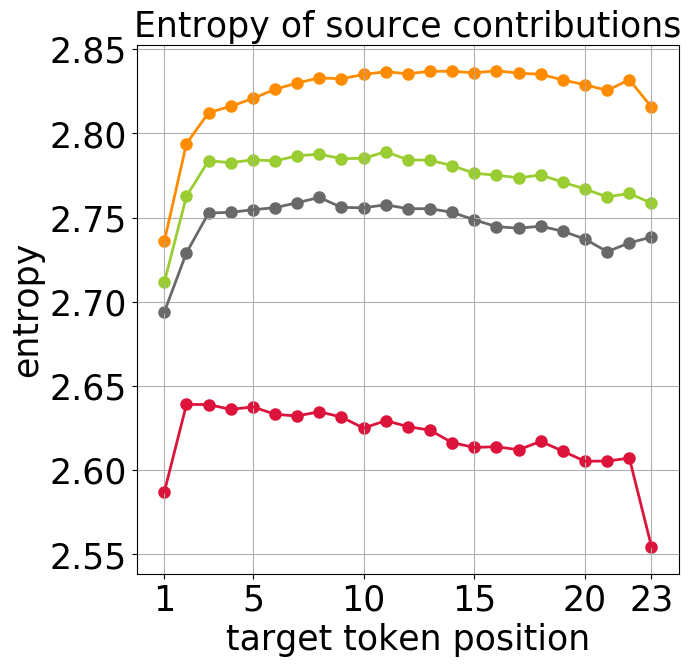}}
    \vspace{-1ex}
  \caption{Contribution of source (a) and entropy of source contributions (b) with random prefixes. %For each generation step, the figure shows (a)~contribution of source, (b)~entropy of source contributions.
  }
  \vspace{-2ex}
    \label{fig:mrt_wd_random_prefix}
\end{figure}

We consider two types of prefixes: model-generated and random. Random prefixes are our main interest here.
We feed prefixes that are fluent but unrelated to the source and look whether a model is likely to fall into a language modeling regime, i.e., to what extent it ignores the source. For model-generated prefixes, we do not expect to see large differences in contributions: this mode is `easy' for the model and the source contributions are high (see Section~\ref{sect:getting_acquainted_prefixes}). The results are shown in Figures~\ref{fig:mrt_wd_model_prefix} and ~\ref{fig:mrt_wd_random_prefix}.

\paragraph{Model-generated prefixes.}  
We see that both MRT and target-side word dropout increase influence or the source, while source-side word dropout does not lead to noticeable changes in the total source contribution (Figure~\ref{fig:mrt_wd_model_prefix}). This agrees with our hypothesis that exposure bias leads to higher reliance on the target. Results for the entropy of source contributions show that only MRT makes the model more confident in the choice of relevant source tokens, while both variants of word dropout force it to rely on broader context (Figure~\ref{fig:mrt_wd_model_prefix}b).

%MRT causes more prominent changes in contributions (Figure~\ref{fig:mrt_wd_model_prefix}). We see the largest difference in the beginning and the end of the generation process, which may be expected when comparing models trained with token-level and sequence-level objectives. The direction of change, i.e.\ decreasing influence of source, is rather unexpected; we leave a detailed investigation of this behavior to future work. For word dropout, changes in the amount of contributions are less noticeable; we see, however, that target-side word dropout makes the model more confident in the choice of relevant source tokens (Figure~\ref{fig:mrt_wd_model_prefix}b).  

% \begin{figure}[t!]
%     \centering
%     \subfloat[]
%     {\includegraphics[scale=0.20]{pict/model_trs_source_mrt_wd.png}}
%     \quad
%     \subfloat[]
%     {\includegraphics[scale=0.20]{pict/model_trs_entropy_mrt_wd.png}}
%     \vspace{-1ex}
%     \subfloat[]
%     {\includegraphics[scale=0.20]{pict/random_prefix_source_mrt_wd.png}}
%     \quad
%     \subfloat[]
%     {\includegraphics[scale=0.20]{pict/random_prefix_entropy_mrt_wd.png}}
%     \vspace{-1ex}
%   \caption{Results for model-generated prefixes (a,~b) and random prefixes (c,~d). For each generation step, the figure shows (a,~c)~contribution of source, (b,~d)~entropy of source contributions.}
%   \vspace{-2ex}
%     \label{fig:mrt_wd}
% \end{figure}

\paragraph{Random prefixes.}  We see that, among all models, the MRT model has the highest influence of source (Figure~\ref{fig:mrt_wd_random_prefix}a) and the most focused source contributions (Figure~\ref{fig:mrt_wd_random_prefix}b).
This agrees with our expectations: by construction, MRT removes exposure bias completely. Therefore, it is confused by random prefixes less than other models.
Additionally, this also links to~\citet{wang-sennrich-2020-exposure} who showed that MRT reduces hallucinations.

When using word dropout, the target-side variant increases the influence of source (i.e. decreases the influence of target), while the source-side variant decreases (Figure~\ref{fig:mrt_wd_random_prefix}a). This is expected: replacing some words with random decreases model's reliance on the corresponding part of input (either the source or the prefix). Note also that target-side word dropout slightly reduces exposure bias (in contrast to source-side word dropout), which gives us another piece of evidence that reducing exposure bias increases the influence of the source.

%When using word dropout, both its variants also increase the influence of source, but to a much lesser extent (Figure~\ref{fig:mrt_wd_random_prefix}a). As expected, since target-side word dropout slightly reduces exposure bias (in contrast to source-side word dropout), it leads to a larger increase of source influence.

Experiments in this section highlight that the methodology we propose can be applied to study exposure bias, robustness, and hallucinations, both in machine translation and more broadly for other language generation tasks. In this work, however, we want to
illustrate more broadly the potential of this approach. In the following, we will compare models trained with varying amounts of data and will look into the training process.

\section{Data Amount}
\label{sect:data_amount}

In this section, we show how the results from Section~\ref{sect:getting_acquainted} change when increasing the amount of training data. 
The observed patterns are the same when evaluating on datasets with reference translations or the ones generated by the corresponding model (in each case, all sentences in the evaluation dataset have the same length). In the main text, we show figures for references. 

\begin{figure}[t!]
    \centering
    {\includegraphics[scale=0.28]{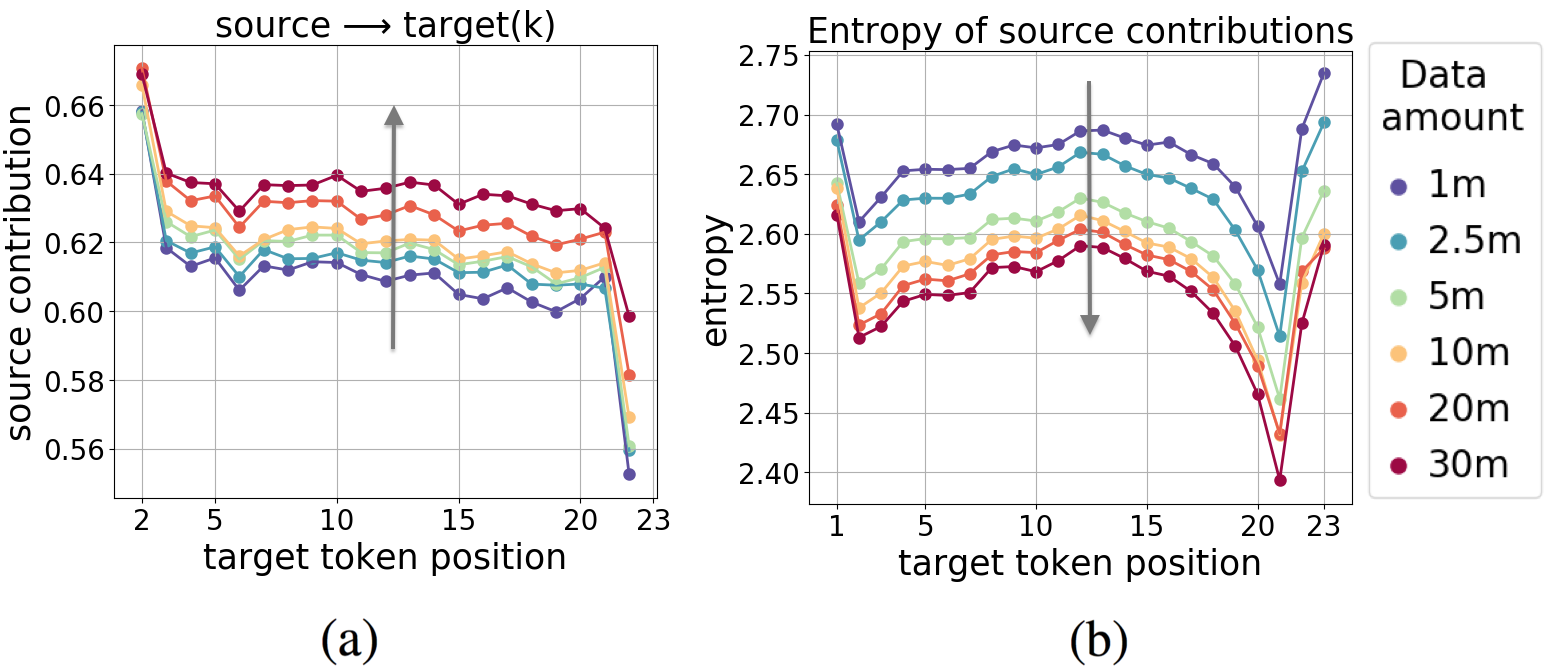}}
    \vspace{-1ex}
  \caption{(a)~source contribution, (b)~entropy of source contributions. The arrows show the direction of change when increasing data amount. (For clarity, in (a) the last two positions (punct. and  EOS) are not shown).}
  \vspace{-2ex}
    \label{fig:data_graphs}
\end{figure}

\paragraph{More data $\Longrightarrow$ higher source contribution.} Figure~\ref{fig:data_graphs}(a) shows the source contribution at each generation step. We can see that, generally, models trained with more data rely on source more heavily.

\paragraph{More data $\Longrightarrow$ more focused contributions.} Figure~\ref{fig:data_graphs}(b) shows that at each generation step, entropy of source contributions decreases with more data. This means that with more training data, the model becomes more confident in the choice of important tokens. 

% \begin{figure}[t!]
%     \centering
%     \subfloat[]
%     {\includegraphics[scale=0.35]{pict/data_src_contribution_with_arrows.png}}\vspace{-1ex}\\
%     \subfloat[]
%     {\includegraphics[scale=0.25]{pict/data_entropy_of_src_with_arrows.png}}
%     \vspace{-1ex}
%   \caption{At each generation step, the figure shows (a)~source contribution, (b)~entropy of source contributions. The arrows show the direction of change when increasing the amount of  data. (For clarity, in (a) the last two positions, punctuation and  EOS, are not shown). }
%   \vspace{-2ex}
%     \label{fig:data_graphs_src}
% \end{figure}

% \begin{figure}[t!]
%     \centering
%     {\includegraphics[scale=0.25]{pict/data_entropy_of_src_with_arrows.png}}
%   \caption{For each generation step, the figure shows entropy of source contributions. The arrows show the direction of change when increasing the amount of data.}
%   \vspace{-2ex}
%     \label{fig:data_graphs_entropy}
% \end{figure}

\begin{figure*}[t!]
    \centering
    {\includegraphics[scale=0.6]{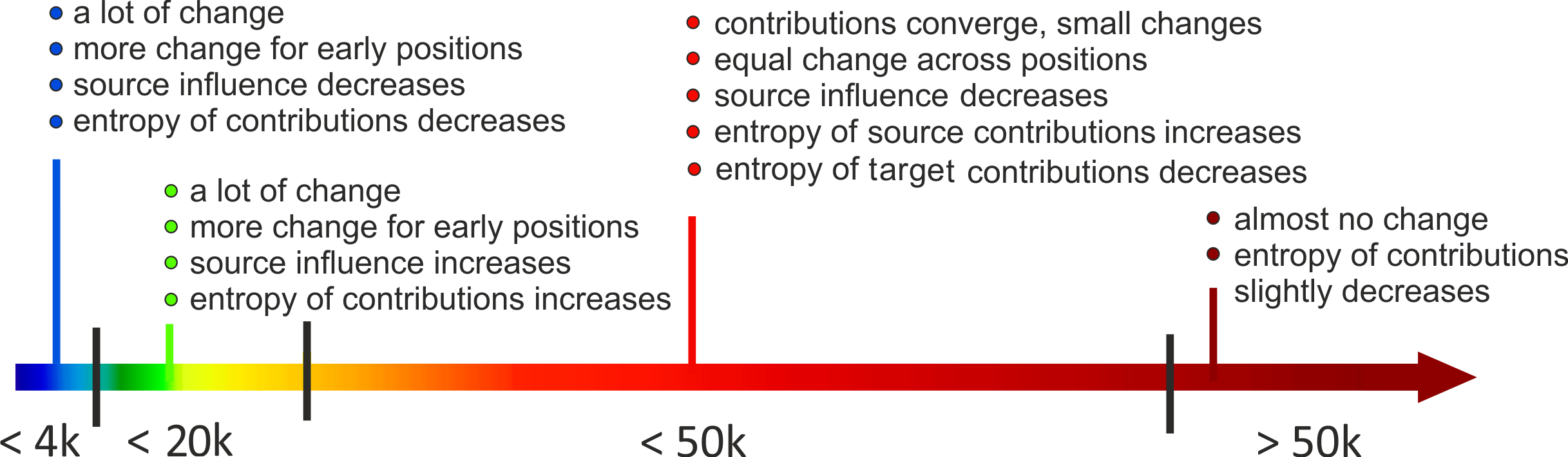}}
\vspace{-1ex}
  \caption{Training timeline.}
  %\vspace{-2ex}
    \label{fig:training_timeline}
\end{figure*}

\section{Training Stages}
\label{sect:training_stages}

\begin{figure*}[t!]
    \centering
    \subfloat[]
    {\includegraphics[scale=0.21]{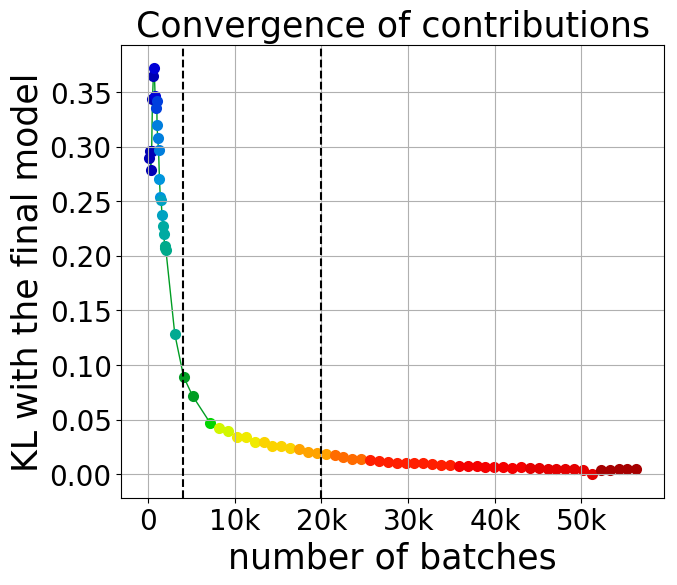}}
    \quad
    \subfloat[]
    {\includegraphics[scale=0.21]{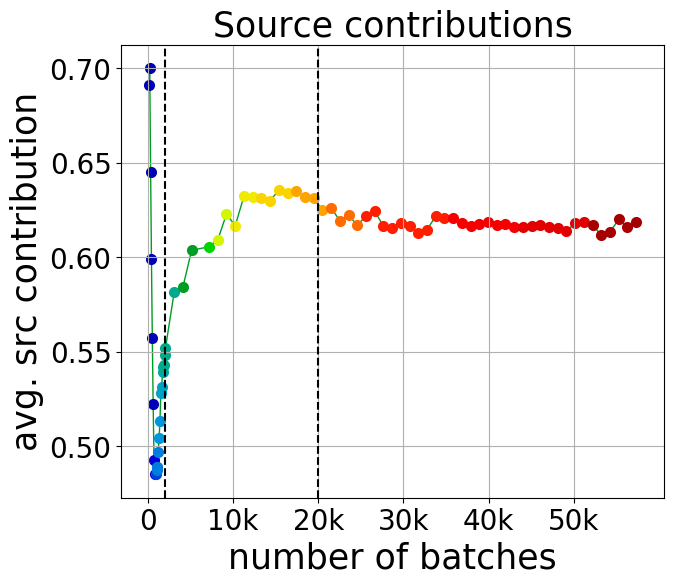}}
    \quad
    \subfloat[]
    {\includegraphics[scale=0.21]{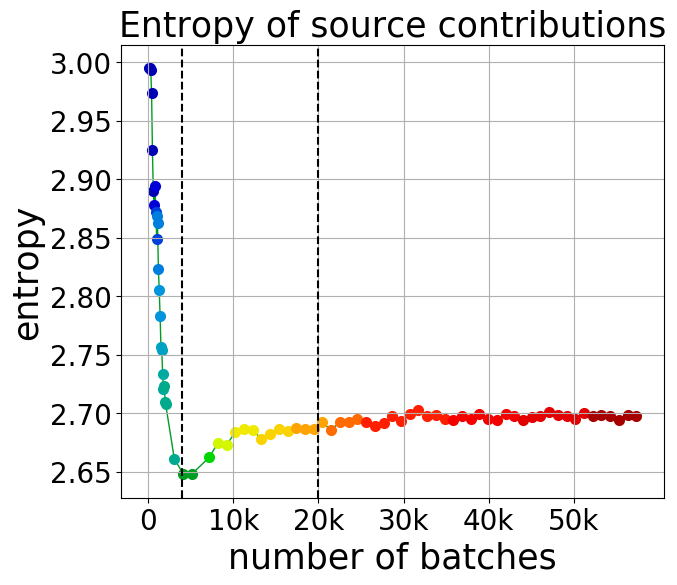}}
    \quad
    \subfloat[]
    {\includegraphics[scale=0.21]{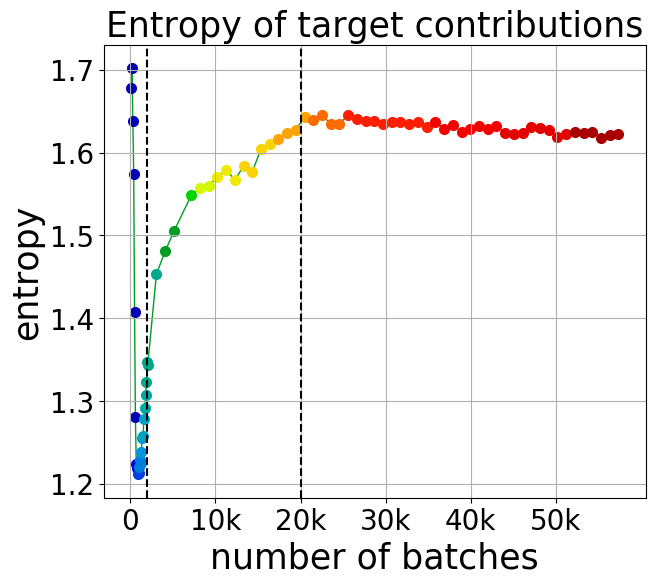}}
    \vspace{-1ex}
  \caption{Training process: (a)~convergence of contributions, (b)~source contribution, (c-d)~entropy of source and target contributions. The model trained on 1m subsample of WMT14 En-Fr dataset. The results are averaged over target positions and evaluation examples.}
  \vspace{-2ex}
    \label{fig:training_convergence}
\end{figure*}

Now we turn to analyzing the training process of an NMT model. 
Specifically, we look at the changes in how the predictions are formed: changes in the amount of source/target contributions and in the entropy of these contributions. Our findings are summarized in Figure~\ref{fig:training_timeline}. In the following, we explain them in more detail. In Section~\ref{sect:training_related}, we draw connections between our training stages (shown in Figure~\ref{fig:training_timeline}) and the ones found in previous work focused on
validating the lottery ticket hypothesis.

\paragraph{Contributions converge early.} First, we evaluate how fast the contributions converge, i.e., how quickly a model understands which tokens are the most important for prediction. For this, 
at each generation step $t$ we evaluate the KL divergence in token influence distributions $(\textsc{r}_t(x_1), \dots, \textsc{r}_t(x_S), \textsc{r}_t(y_1), \dots, \textsc{r}_t(y_{t-1}))$ from the final converged model to the model in training.
Figure~\ref{fig:training_convergence}(a) shows that contributions converge early. After approximately 20k batches, the model is very close to its final state in the choice of tokens to rely on for a prediction.

\paragraph{Changes in training are not monotonic.} Figures~\ref{fig:training_convergence}(b-d) show how the amount of source contribution and the entropy of source and target contributions change in training. We see that all three figures have the same distinct stages (shown with vertical lines). First, source influence decreases, and both source and target contributions become more focused. In this stage, most of the change happens  (Figure~\ref{fig:training_convergence}(a)). In the second stage, the model also undergoes substantial change, but all processes change their direction: source influence increases and the model learns to rely on broader context (entropy is increasing). Finally, in the third stage, the direction changes again for the total source contribution and the entropy of target contributions, and remains the same for the entropy of source contributions. However, very little is going on -- the model slowly converges.

These three stages correspond to the first three stages shown in Figure~\ref{fig:training_timeline}; at this point, the model trained on 1m sentence pairs converges. With more data (e.g., 20m sentence pairs), we further observed the next stage (the last one in Figure~\ref{fig:training_timeline}), where the entropy of both source and target contributions is decreasing again. However, this last stage is much slower than the third, and the final state does not differ much from the end of the third stage.

\begin{figure}[t!]
    \centering
    % \subfloat[Source contribution at each target position]
    % {\includegraphics[scale=0.3]{pict/ckpts_src_infl_by_dst_pos_full_v4_with_arrows.png}}
    %  \\
    %  \vspace{-1ex}
    %\subfloat[Entropy of source contributions]
    {\includegraphics[scale=0.3]{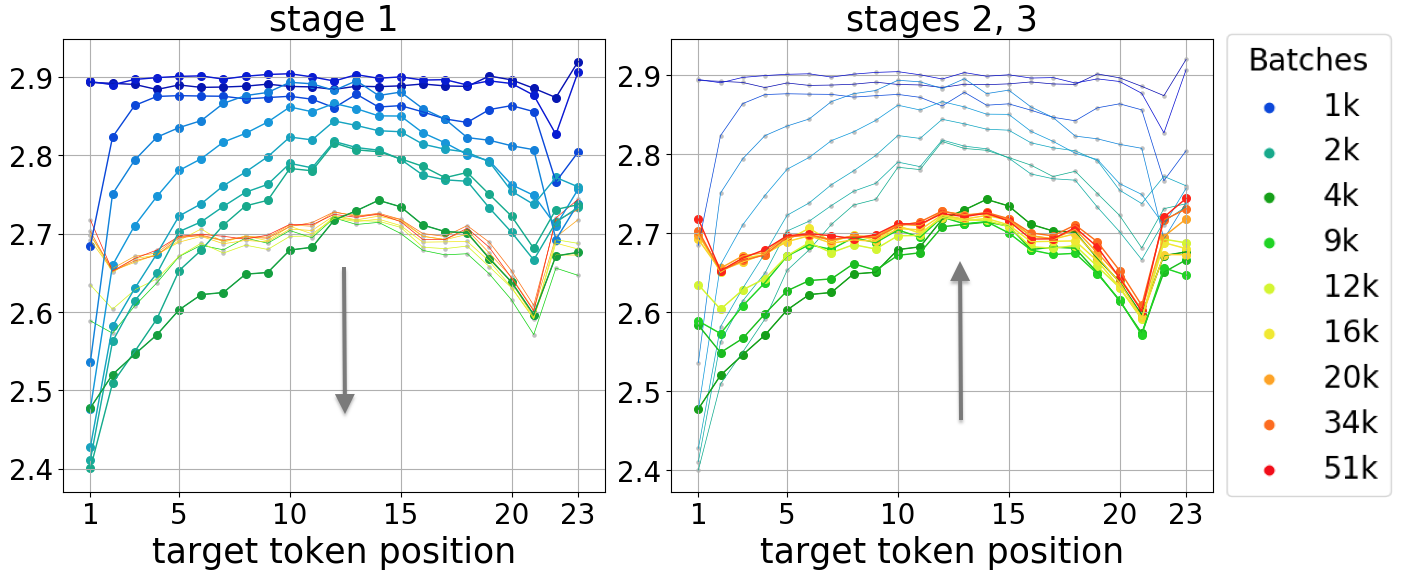}}

  \caption{Entropy of source contributions. Changes in training for each target position; each line corresponds to a model state. The arrows show the direction of change when the training progresses. In the figures, all stages are shown, but the stages of interest are highlighted more prominently.}
  \vspace{-2ex}
    \label{fig:training_by_target_position}
\end{figure}

\begin{figure}[t!]
    {\includegraphics[scale=0.3]{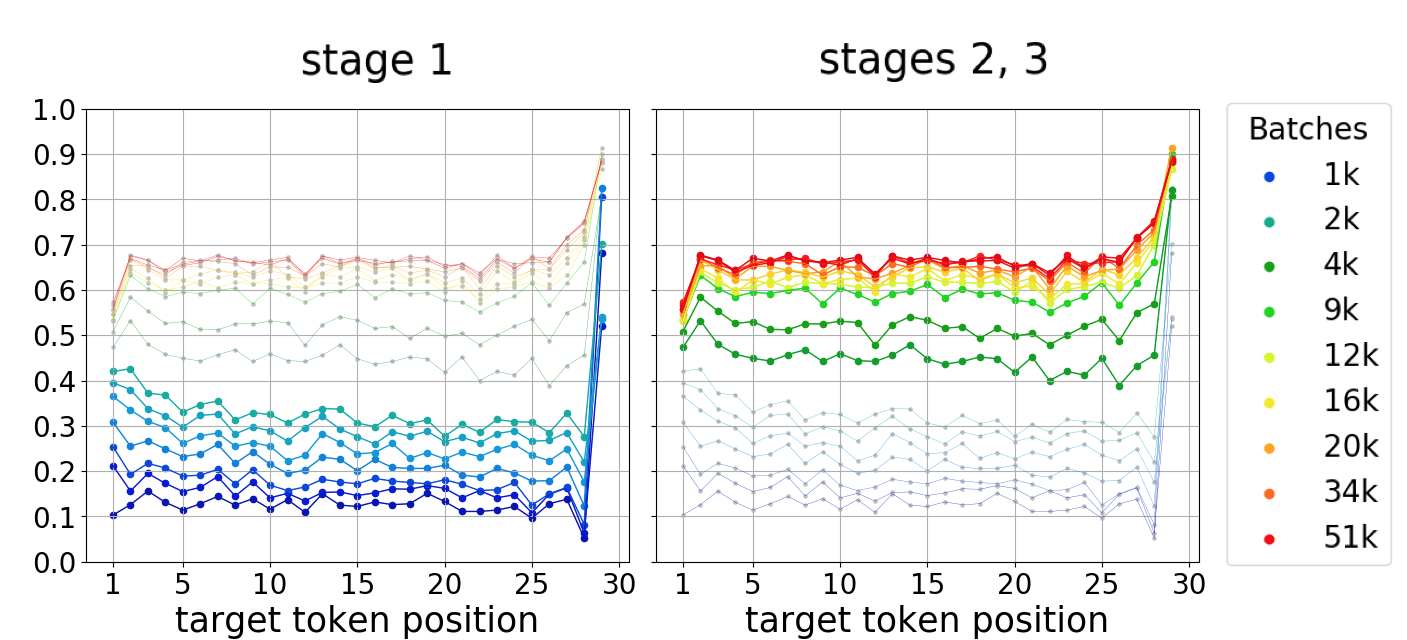}}
  \caption{Accuracy change for each target position; each line corresponds to a model state. In the figures, all stages are shown, but the stages of interest are highlighted more prominently.}
  \vspace{-2ex}
    \label{fig:training_accuracy}
\end{figure}

\paragraph{Early positions change more.} Figure~\ref{fig:training_by_target_position} shows how entropy of source contributions changes for each target position. We see that earlier positions are the ones that change most actively: at these positions, we see the largest decrease at the first stage and the largest following increase at the subsequent stages. If we look at how accuracy for each position changes in training (Figure~\ref{fig:training_accuracy}), we see that at the end of the first stage, early tokens have the highest accuracy.\footnote{Accuracy is the proportion of cases where the correct token is the most probable choice.} This is not surprising: one could expect early positions to train faster because they are observed more frequently in training. Previously such intuition motivated the usage of sentence length as one of the criteria for curriculum learning (e.g., \citet{kocmi-bojar-2017-curriculum}).

% \begin{figure}[t!]
%     \centering
%      \subfloat[Entropy of source contributions]
%     % {\includegraphics[scale=0.3]{pict/ckpts_src_infl_by_dst_pos_full_v4_with_arrows.png}}
%     %  \\
%     %  \vspace{-1ex}
%     %\subfloat[Entropy of source contributions]
%     {\includegraphics[scale=0.3]{pict/ckpts_src_entropy_full_v4_with_arrows.png}}
% \vspace{-2ex}
% \subfloat[Token-level accuracy]
% {\includegraphics[scale=0.3]{pict/ckpts_acc_by_dst_pos_full_v4.png}}

%   \caption{Changes in training for each target position; each line corresponds to a model state. The arrows show the direction of change when the training progresses. In the figures, all stages are shown, but the stages of interest are highlighted more prominently.}
%   \vspace{-2ex}
%     \label{fig:training_by_target_position}
% \end{figure}

\subsection{Relation to Previous Work}
\label{sect:training_related}

Interestingly, our stages in Figure~\ref{fig:training_timeline} agree with the ones found by~\citet{Frankle2020The_early} for ResNet-20 trained on CIFAR-10 when investigating, among other things, the lottery ticket hypothesis~\cite{frankle2018the_lottery}.
Their stages were defined based on the changes in gradient magnitude, in the weight space, in the performance, and in the effectiveness of rewinding in search of the `winning' subnetwork (for more details on the lottery ticket hypothesis and the rewinding, see the work by~\citet{frankle2019stabilizing}). 
Comparing the stages by \citet{Frankle2020The_early} with ours, we see that (1)~their relative sizes in the corresponding timelines match well, (2)~the rewinding starts to be effective at the third stage; for our model, this is when the contributions have almost converged. In future work, it would be interesting to further investigate this relation.

% \begin{figure}[t!]
%     {\includegraphics[scale=0.3]{pict/ckpts_acc_by_dst_pos_full_v4.png}}
%   \caption{Accuracy change for each target position; each line corresponds to a model state. In the figures, all stages are shown, but the stages of interest are highlighted more prominently.}
%   \vspace{-2ex}
%     \label{fig:training_accuracy}
% \end{figure}

\section{Additional Related Work}

To estimate the influence of source to an NMT prediction, \citet{ma2018analysis} trained an NMT model with an auxiliary second decoder where the encoder context vector was masked. Then the source influence was measured as the KL divergence between predictions of the two decoders. However, the ability of an auxiliary decoder to generate similar distribution is not equivalent to the main model not using source. More recently, as a measure of individual token importance, \citet{he-etal-2019-towards} used Integrated Gradients~\cite{pmlr-v70-sundararajan17a}.

In machine translation, LRP was previously used for visualization~\cite{lrp-ding-2017} and to find the most important attention heads in the Transformer's encoder~\cite{voita-etal-2019-analyzing}. Similar to our work, \citet{voita-etal-2019-analyzing} evaluated LRP on average over a dataset (and not for a single prediction) to extract patterns in model behaviour. Both works used the more popular $\varepsilon$-LRP, while for our analysis, the $\alpha\beta$-LRP was more suitable (Section~\ref{sect:lrp}). %because it defines relevances at each step in such a way that they are positive. L: to win space
For language modeling, \citet{calvillo-crocker-2018-language} use LRP to evaluate relevance of neurons in RNNs for a small synthetic setting.

\section{Conclusions}

We show how to use LRP to evaluate the relative contributions of source and target to NMT predictions. We illustrate the potential of this approach by analyzing changes in these contributions when conditioning on different types of prefixes (references, model predictions or random translations), when varying training objectives or the amount of training data, and during the training process. Some of our findings are: (1)~models trained with more data rely on source more and have more sharp token contributions; (2)~the training process is non-monotonic with several distinct stages. These stages agree with the ones found in previous work focused on validating the lottery ticket hypothesis, which suggests future investigation of this connection. 
Additionally, we show that models suffering from exposure bias are more prone to over-relying on target history (and hence to hallucinating) than the ones where the exposure bias is mitigated.
In future work, our methodology can be used to measure the effects of different and novel training regimes on the balance of source and target contributions.

\section*{Acknowledgments} We would like to thank the anonymous reviewers for their comments. We also thank Wenxu Li for noticing an error in the first version of our released code. The work is partially supported by the European Research Council (Titov, ERC StG BroadSem 678254), Dutch NWO (Titov, VIDI 639.022.518) and EU Horizon 2020 (GoURMET, no. 825299). Lena is supported by the Facebook PhD Fellowship. Rico Sennrich acknowledges support of the Swiss National Science Foundation (MUTAMUR; no. 176727). 

\bibliography{acl2021}
\bibliographystyle{acl_natbib}

\newpage
\phantom{0}
\newpage
\appendix

\section{Experimental setup}

\subsection{Data preprocessing}

We use random subsets of the WMT14 En-Fr dataset: \url{http://www.statmt.org/wmt14/translation-task.html}. Sentences were encoded using byte-pair encoding~\cite{sennrich-bpe}, with source and target vocabularies of about 32000 tokens. 
Translation pairs were batched together by approximate sequence length. Each training batch contained a set of translation pairs containing approximately 16000\footnote{This can be reached by using several of GPUs or by accumulating the gradients for several batches and then making an update.} source tokens for 1m subsample and 32000 for larger datasets.

\subsection{Model parameters}

We follow the setup of Transformer base model~\cite{attention-is-all-you-need}. More precisely, the number of layers in the encoder and in the decoder is $N=6$. We employ $h = 8$ parallel attention layers, or heads. The dimensionality of input and output is $d_{model} = 512$, and the inner-layer of a feed-forward networks has dimensionality $d_{ff}=2048$.

We use regularization as described in~\cite{attention-is-all-you-need}.

\subsection{Optimizer}
The optimizer we use is the same as in~\cite{attention-is-all-you-need}.
We use the Adam optimizer~\cite{adam-optimizer} with $\beta_1 = 0{.}9$, $\beta_2 = 0{.}98$ and $\varepsilon = 10^{-9}$. We vary the learning rate over the course of training, according to the formula:
\begin{multline*}
l_{rate}=scale\cdot \min(step\_num^{-0.5},\\ step\_num\cdot warmup\_steps^{-1.5}) 
\end{multline*}
We use $warmup\_steps = 16000$, $scale=4$.

We train models till convergence and average 5 latest checkpoints. Approximate number of training batches are: 57k for 1m dataset, 220k for 2.5m dataset and 600k for the rest.

\section{Minimum Risk Training}

\subsection{Background}

Minimum Risk Training (MRT) minimises the expected loss (`risk') with respect to the posterior distribution:
\begin{equation*}
\mathcal{R}(\theta) = \sum\limits_{(x, y)}\sum\limits_{\Tilde{y}\in\mathcal{Y}(x)}P(\Tilde{y}|x, \theta)\Delta(\Tilde{y}, y),
\end{equation*}
where $\mathcal{Y}(x)$ is a set of all possible candidate translations for $x$,  $\Delta(\Tilde{y}, y)$ is the discrepancy between the model prediction $\Tilde{y}$ and the gold translation $y$. 

Since the search space $\mathcal{Y}(x)$ is exponential, 
in practice it is common to use only a subset of the full space. Formally, instead of $\mathcal{Y}(x)$ we use $\mathcal{S}(x)\in \mathcal{Y}(x)$, where $\mathcal{S}(x)$ is obtained by sampling several translations. The probabilities $P(\Tilde{y}|x, \theta)$ are replaced with the $\Tilde{P}$, which is renormalized over the subset $\mathcal{S}$:
\begin{equation*}
\Tilde{P}(\Tilde{y}|x, \theta, \alpha) = \frac{P(\Tilde{y}|x, \theta)^{\alpha}}{\sum\limits_{y'\in\mathcal{S}(x)}P(y'|x, \theta)^{\alpha}}.
\end{equation*}

The hyperparameter $\alpha$ is used to control the sharpness of the distribution.

\subsection{Experimental setting}

To choose the setting, we mostly relied on previous work~\cite{shen-etal-2016-minimum,edunov-etal-2018-classical}. Model is pre-trained with the token-level objective MLE and then fine-tuned with MRT; the fine-tuning stage is approximately one epoch.

\paragraph{Candidate translations.} The translations are sampled using standard random sampling without temperature. Following~\citet{shen-etal-2016-minimum}, we take the large number of candidates; specifically, we use 50 translations and add a reference to the subset. While \citet{edunov-etal-2018-classical} report that adding the reference to the set of candidates hurts quality, in preliminary experiments we found that this was not the case for our setting.

\paragraph{Measure of discrepancy.} The measure of discrepancy, $\Delta(\Tilde{y}, y)$, is a negative smoothed sentence-level BLEU. 

\paragraph{Batch size.} On average, the number of examples (where an example is a translation pair along with all candidates) is the same as in training of the baseline models. This is achieved by accumulating gradients for several steps and making an update.

\paragraph{Other parameters.} Following \cite{wang-sennrich-2020-exposure}, we set $\alpha=0.005$ and the learning rate to $0{.}00001$.

\end{document}